\pdfoutput=1
\documentclass[11pt]{article}

\usepackage[preprint]{acl}

\usepackage{times}
\usepackage{latexsym}

\usepackage[T1]{fontenc}
\usepackage[utf8]{inputenc}

\usepackage{microtype}

\usepackage{inconsolata}

\usepackage{graphicx}
\usepackage{graphics}

\usepackage{array}           
\usepackage{tikz}            
\usepackage{amssymb}         
\usepackage{xcolor}          
\usepackage{colortbl}        
\usepackage{makecell}        
\usepackage{caption}         
\usepackage{multirow}
\usepackage{verbatim}
\usepackage{verbatimbox}
\usepackage{float}

\title{Fact or Fiction? Improving Fact Verification with Knowledge Graphs through Simplified Subgraph Retrievals}

\author{Tobias A. Opsahl \\
  University of Oslo  \\
  \texttt{tobiasao@uio.no}
  }

\begin{document}
\maketitle
\begin{abstract}

Despite recent success in natural language processing (NLP), fact verification still remains a difficult task. Due to misinformation spreading increasingly fast, attention has been directed towards automatically verifying the correctness of claims. In the domain of NLP, this is usually done by training supervised machine learning models to verify claims by utilizing evidence from trustworthy corpora. We present efficient methods for verifying claims on a dataset where the evidence is in the form of structured knowledge graphs. We use the \textsc{FactKG} dataset, which is constructed from the \emph{DBpedia} knowledge graph extracted from Wikipedia. By simplifying the evidence retrieval process, from fine-tuned language models to simple logical retrievals, we are able to construct models that both require less computational resources and achieve better test-set accuracy.

\end{abstract}

\section{Introduction}

As the volume of information generated continues to grow, so does the risk of misinformation spreading, which has made automatic fact verification a crucial task in NLP \citep{cohen2011computational_journalism, hassan2015quest_to_automate_fact-checking, thorne2018automated_fact_checking, bekoulis2021review_on_fact_extraction}. Traditionally, fact verification has been tackled in journalism by experts manually researching topics and writing articles about their findings. Some specific websites dedicated to this approach are \emph{FactCheck.org} and \emph{PolitiFact.com}. However, it is time consuming and labor intensive, and is not able to follow the pace of the creation of information in digital media \citep{cohen2011computational_journalism, hassan2015quest_to_automate_fact-checking}.

One of the most popular datasets for fact verification is the \emph{Fact Extraction and VERification} (FEVER) dataset \citep{thorne2018fever}. It consists of claims supported by a corpus of Wikipedia articles. Models trained on the dataset need to extract the relevant evidence and use it to classify claims as \emph{supported}, \emph{refuted} or \emph{not enough information}.

Despite its popularity, several issues have been discovered. Due to the manual construction of claims, the structure of the language is inherently biased with respect to the classes, and therefore it is possible to achieve good performance without using the evidence at all \citep{schuster2019towards_debiasing}. It has also been shown that models trained on FEVER experience a significant drop in performance when the factual evidence is changed in a way that influences the validity of claims \citep{hidey2020deseption}. These issues can be improved by accordingly adjusting the validation and test dataset to contain less biased data \citep{schuster2019towards_debiasing, hidey2020deseption}, but we believe it is important to develop models on other datasets as well.

A less studied approach to process evidence is by structured data. In many real-world examples, data is available in large structured databases, rather than unstructured articles. This is relevant for domains such as social networks, logistics, management systems and database systems. The dataset \emph{TabFact} \citep{chen2019tabfact} is created with this intent, consisting of claims with tabular evidence extracted from Wikipedia. 

\begin{figure}[tb]
    \begin{center}
    \includegraphics[width=1.0\linewidth]{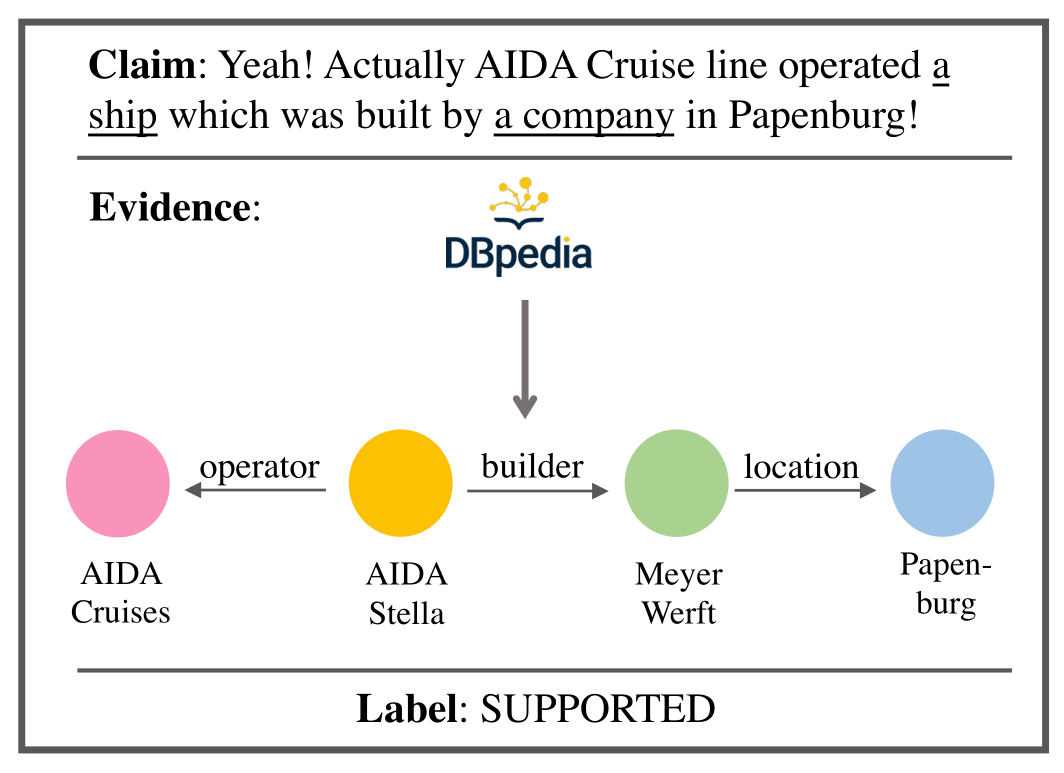}
    \end{center}
    \caption{An example claim from \textsc{FactKG} \citep{kim2023factkg}. The claim can be verified or refuted based on the DBpedia KG \citep{lehmann2015dbpedia}. This is Figure 1 from \citet{kim2023factkg}.}
    \label{fig:dbpedia}
    \vspace{-3mm}
\end{figure}

We will dedicate this article to increase the performance of models trained on the \textsc{FactKG} dataset \citep{kim2023factkg}, a dataset created for fact verification with structured evidence in the form of \emph{knowledge graphs} (KGs). The claims are created with evidence from \emph{DBpedia} \citep{lehmann2015dbpedia}, a large KG extracted from Wikipedia. A KG consists of nodes and edges linked together to represent structural concepts. Nodes represent entities, such as persons, things or events, and edges represent relations, conveying how entities are related, as shown in Figure~\ref{fig:dbpedia}. For instance, a node can be the company \emph{Meyer Werft}, and since it is located in the city \emph{Papenburg}, they are connected with the edge \emph{location}. We refer to \textit{Meyer Werft, location, Papenburg} as a \emph{knowledge triple}.

Since the task of fact verification with KGs remains relatively unexamined, we want to explore several different approaches to the problem. We use the following three model architectures:

\begin{itemize}
    \item \textbf{Textual Fine-tuning:} Fine-tuning pretrained encoder models on text data for claim verification. We use BERT \citep{devlin2018bert} by concatenating the claims with subgraphs represented as strings.

    \item \textbf{Hybrid Graph-Language Model:} Using a modification of a \emph{question answer graph neural network} (QA-GNN) \citep{yasunaga2021qa_gnn}, which both uses a pretrained encoder model to embed the claim, and a graph neural network (GNN) to structurally process the subgraphs.

    \item \textbf{LLM Prompting:} Deploying state-of-the-art language models in a few-shot setting, without the need for additional finetuning. We use ChatGPT 4o \citep{achiam2023gpt4, 2024chatgpt4o} for this setting.
\end{itemize}

The textual finetuning serves as a simple and conventional method, while the QA-GNN can handle graph based data efficiently and is more specifically constructed for the task of interest. In contrast, the LLM prompting displays how well general purpose language models can perform on the task. It does not require any further training and does not use any evidence. Therefore, it will serve as a baseline and give insight to how difficult the task is.

Our main contribution is that we increase the accuracy and computational efficiency of models trained on \textsc{FactKG}. By utilizing efficient subgraph retrieval methods, we are able to substantially increase the test-set accuracy from 77.65\% \citep{kim2023factkg} to 93.49\%. To the best of the authors knowledge, this is the best performance achieved so far on the dataset. Additionally, our models train quicker, taking only 1.5-10 hours, compared to the 2-3 days spent on the benchmark model from \citet{kim2023factkg}, reported by the authors. The code and documentation can be found at \url{https://github.com/Tobias-Opsahl/Fact-or-Fiction}.

\section{Related Work}

\subsection{Fact Verification}

The FEVER dataset is one of the most popular datasets used for fact verification \citep{thorne2018fever}, and has influenced several model architectures. \emph{Graph-based Evidence Aggregating and Reasoning} (GEAR) \citep{zhou2019gear} works by finding relevant articles with entity linking, giving them a relevance score, embedding the claim and sentences in the relevant evidence with a pre-trained BERT \citep{devlin2018bert}, and then using a GNN to reason over the embeddings. The \emph{Neural Semantic Matching Network} (NSMN) \citep{nie2019combining} used three homogenous neural networks used for document retrieval, sentence selection and claim verification. By using a transformer based architecture, \emph{Generative Evidence REtrieval} (GERE) \citep{chen2022gere} combined the evidence retrieval and sentence identifying into a single step.

Several other datasets for fact verification have also been proposed. The \emph{Fake News Challenge} \citep{hanselowski2018retrospective_fake_news} were aimed towards predicting the relevance and agreement of a title and text. \emph{VitaminC} \citep{schuster2021get_your_vitamin_C} focuses on representing changing evidence, and was created by constructing claims based on different revisions of Wikipedia articles. The dataset \emph{FAVIQ} \citep{park2021faviq} explored ambiguous parts of claims, while \emph{TabFact} \citep{chen2019tabfact} used tabular data as evidence. There have also been proposed multimodular dataset for fact verification, combining claims and images \citep{zlatkova2019fact-checking_meets_fauxtography, mishra2022factify}.

\subsection{The FactKG Dataset}

The \textsc{FactKG} dataset \citep{kim2023factkg} consists of 108 000 English claims for fact verification, where the downstream task is to predict whether the claims are true or false. The claims are constructed from the DBpedia KG \citep{lehmann2015dbpedia}, which is extracted from Wikipedia and represents how entities are related to each other.

The claims are constructed on either of the following five reasoning types:

\begin{itemize}
    \item \textbf{One-hop:} To answer a one-hop claim, one only needs to traverse one edge in the KG. In other words, only one knowledge triple is needed to verify the validity of the claim.
    \item \textbf{Multi-hop:} As opposed to one-hop claims, one needs to traverse multiple steps in the KG to verify multi-hop claims.
    \item \textbf{Conjunction:} The claim includes a combination of multiple claims, which are often added together with the word \emph{and}.
    \item \textbf{Existence:} These claims state that an entity has a relation, but does not specify which entity it relates to.
    \item \textbf{Negation:} The claim contains negations, such as \emph{not}.
\end{itemize}

The dataset is split in a train-validation-test set of proportion 8:1:1. The train and validation set includes relevant subgraphs for each claim, but not the test set. All claims include a list of entities present in the claim and as nodes in the KG.

\subsection{Question Answer Graph Neural Networks}

The \emph{question answer graph neural network} (QA-GNN) \citep{yasunaga2021qa_gnn} is a hybrid language and GNN model that both uses a pre-trained language model to process the text, and couples it with a GNN reasoning over a subgraph. It is given text and a subgraph as input. The text, consisting of a question and possible answers, is added as a node to the subgraph. The language model embeds the text, and assigns a relevance score to each node in the subgraph. The relevance scores are multiplied with the node features, before being sent into the GNN. The GNN output, text-node and the text embedding are concatenated before being put into the classification layer.

\section{Methods}

\subsection{Efficient Subgraph Retrieval}

We experiment with different ways of retrieving relevant subgraphs for the claim, focusing on computational efficiency. Each datapoint in the \textsc{FactKG} dataset consists of a claim and a list of entities that appear both in the claim and the KG. Since the part of DBpedia used in FactKG is fairly large (1.53GB), it is necessary to only use a small subgraph of it as input to the models. The benchmark model from \citet{kim2023factkg} uses two language models to predict the relevant edges and the depth of the graph. We wish to simplify this step in order to reduce the model complexity, and propose non-trainable methods for subgraph retrieval.

\begin{figure}[tb]
    \vspace{-2mm}

    \centering
    \includegraphics[width=0.47\textwidth]{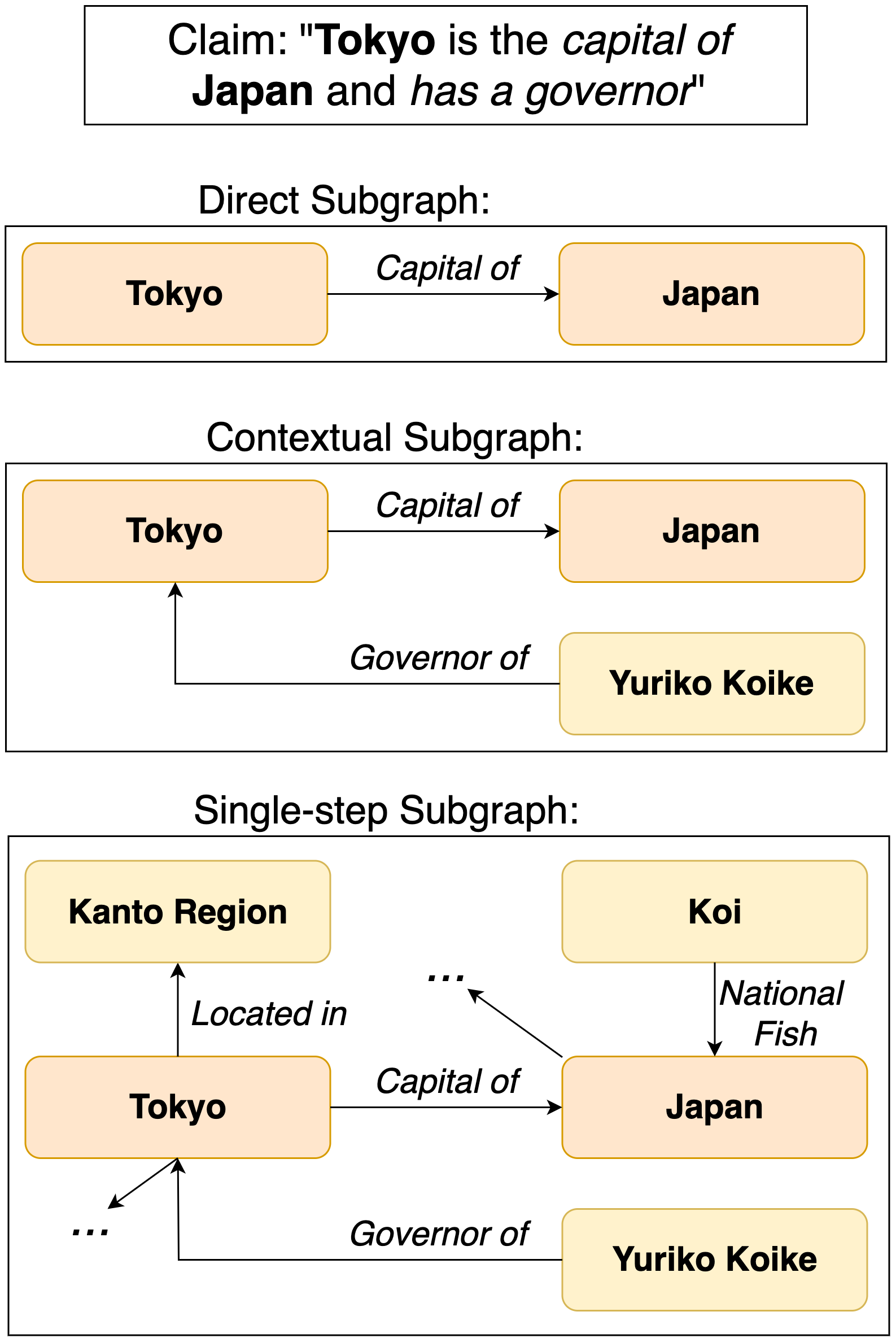}

    \caption{\textbf{Examples of the different subgraphs explored in this article}. Boxes and \textbf{bold letters} represent entities, while arrows and \textit{italic letters} represent relations. This claim is meant for illustrative purposes and is not present in the \textsc{FactKG} dataset.}

\label{fig:subgraph_types}
\end{figure}

We experiment with the following methods (examples can be found in Figure~\ref{fig:subgraph_types}):

\begin{itemize}
    \item \textbf{Direct:} Only includes knowledge triples where both nodes are present in the entity list.
    \item \textbf{Contextualized:} First, includes all direct subgraphs. Additionally, lemmatize the words in the claim and check if the nodes in the entity list have any relations corresponding to the lemmatized words in the claim. Include all knowledge triples where at least one node is in the entity list and the relation is found in the claim.
    \item \textbf{Single-step:} Includes every knowledge triple one can be traversed in one step from a node in the entity list. In other words, include every knowledge triple that contains at least one node in the entity list.
\end{itemize}

\subsection{Finetuning BERT}

We use BERT \citep{devlin2018bert} as our pre-trained language model. We first train a baseline model using only the claims and no subgraphs, and then with all of the different methods for retrieving subgraphs. The subgraphs are converted to strings, where each knowledge triple is represented with square brackets, and the name of the nodes and edges are the same as they appear in DBpedia. The order of the knowledge triples is determined by the order of the list of entities in the FactKG dataset and the order of the edges in DBpedia. The subgraphs are concatenated after the claims and a `` | '' separation token.

\subsection{QA-GNN Architecture}

In order to adapt the QA-GNN to the fact verification setting, we perform some slight modifications. Because the possible answers are always ``true'' or ``false'', we embed only the claims, instead of the question and answer combination. Additionally, we do not connect the embedded question or claim to the subgraph.

We use a pre-trained BERT \citep{devlin2018bert} as the language model to embed and calculate the relevance scores. In order to reduce the complexity of the model, we use a frozen BERT to calculate the embeddings for the nodes and the edges in the graph. This way, all of the embeddings in the graph can be pre-calculated. We use the last hidden layer representation of the CLS token, which is of length 768. The BERT that calculates the relevance scores and the embedding of the claim is not frozen. The relevance scores are computed as the cosine similarity between the claim embedding and the embedding of the text in the nodes.

We use a graph attention network \citep{velivckovic2017graph_attention} for our GNN. Since the subgraphs are quite shallow, we only use two layers in the GNN, and apply batch norm \citep{ioffe2015batch_norm}. Each layer has 256 features, which is mean-pooled and concatenated with the BERT embedding and sent into the classification layer. We add dropout \citep{srivastava2014dropout} to the GNN layers and the classification layer.

\subsection{ChatGPT Prompting}

We construct a prompt for ChatGPT 4o in order to answer a list of claims as accurately as possible. This is done by creating an initial prompt and validating the results on 100 randomly drawn claims from the validation set, and by trying different configurations of the prompt until we do not get a better validation set accuracy. We then use the best prompt with 100 randomly drawn unseen test-set questions, and attempt to ask 25, 50 and 100 claims at a time, to see if the amount of claims at a time influences the performance. We run the testing three times.

Since we do not have access to vast enough computational resources to run an LLM, this analysis is limited by only using 100 datapoints from the test set. In order to get access to a state-of-the-art LLM, we used the ChatGPT website with a ``ChatGPT Plus'' subscription to perform the prompting. This model is not seeded, so the exact answers are not reproducible, but every prompt and answer are available in the software provided with this article. We used the ChatGPT 4o model 30th of May 2024. Every prompt was performed in the ``temporary chat'' setting, so the model did not have access to the history of previous experiments.

Due to the inability to use the entire test set and the lack of reproducibility, we do not directly compare this experiment to the other models. However, we still believe it serves as a valuable benchmark. Recently, the performance of LLMs has rapidly improved, which suggests that their applications will continue to broaden. Additionally, this approach is not fine-tuned, and may work as an interesting benchmark that can contextualize the results of the other models.

\begin{table*}[tb]
    \centering
    \resizebox{0.99\textwidth}{!}{%
    \begin{tabular}{|c|c|ccccc|c|c}
        \Xhline{1.5pt}
        \textbf{Input Type} & \textbf{Model} & \textbf{One-hop} & \textbf{Conjunction} & \textbf{Existence} & \textbf{Multi-hop} & \textbf{Negation} & \textbf{Total} \\ \hline
        \multirow{2}{*}{\textit{Claim Only}}  
        & \textsc{FactKG} BERT Baseline & 69.64 & 63.31 & 61.84 & 70.06 & 63.62 & 65.20 \\
        & FactGenius RoBERTa Baseline & 71 & 72 & 52 & 74 & 54 & 68 \\
        & BERT (no subgraphs) & 67.71 & 67.48 & 62.51 & 73.28 & 64.23 & 68.99 \\
         \hline
        \multirow{3}{*}{\textit{With Subgraphs}}
        & \textsc{FactKG} GEAR Benchmark & 83.23 & 77.68 & 81.61 & 68.84 & 79.41 & 77.65 \\
        & FactGenius RoBERTa-two-stage & 89 & 85 & 95 & 75 & 87 & 85 \\
        & QA-GNN (single-step) & 79.08 & 74.43 & 83.37 & 74.72 & 79.60 & 78.08 \\
        & BERT (single-step) & \textbf{97.40} & \textbf{97.51} & \textbf{97.31} & \textbf{80.32} & \textbf{92.54} & \textbf{93.49} \\
         \hline
    
        \Xhline{1.5pt}
    \end{tabular}
    }
    \caption{\textbf{Test-set accuracy for the best models from this article and the best benchmark models.} The \textsc{FactKG} models are from \citet{kim2023factkg}, while the FactGenius models are from \citet{gautam2024factgenius}. The fine-tuned BERT model performed the best, while the QA-GNN was the computationally most efficient model.}
    \label{tab:best_test_accuracies}
\end{table*}

\subsection{Benchmark Models}

We will compare the results against the best benchmark models from \citet{kim2023factkg} and the best performing models known to the authors, found in \citet{gautam2024factgenius}. These comparisons include both baselines that use only the claims and models that also incorporate subgraph evidence.

\noindent \textbf{Claim-Only Models:}

\begin{itemize}
    \item  \textbf{FactKG BERT Baseline}: The baseline model from \citet{kim2023factkg} uses a fine-tuned BERT, training only on the claims.
    \item \textbf{RoBERTa Baseline}: Similarly to above, the baseline from \citet{gautam2024factgenius} uses a fine-tuned language model with claims only, but uses RoBERTa \citep{liu2019roberta} as the base model.
\end{itemize}

\noindent \textbf{Models Utilizing Subgraphs:}

\begin{itemize}
    \item \textbf{GEAR-Based Model}: The benchmark model from \citet{kim2023factkg} is inspired by GEAR \citep{zhou2019gear}, but has been adapted to handle graph-based evidence. It uses two fine-tuned language models to retrieve the subgraphs. One of them predicts relevant edges, the other predicts the depth of the subgraph.
    \item \textbf{FactGenius:} This model combines zero-shot LLM prompting with fuzzy text matching on the KG \citep{gautam2024factgenius}. The LLM filters relevant parts of the subgraphs, which are then refined using fuzzy text matching. Finally, a fine-tuned RoBERTa is used to make the downstream prediction.
\end{itemize}

\subsection{Further Experimental Details}

Due to computational constraints, we tuned the hyperparameters one by one, instead of performing a grid search. All the training was performed on the University of Oslo's USIT ML nodes \citep{2023uio_mlnodes}, using an RTX 2080 Ti GPU with 11GB VRAM. The BERT model has 109 483 778 parameters, which all were fine-tuned. The QA-GNN used a total of 109 746 945 parameters. The \textsc{FactKG} dataset comes with a lighter version of DBpedia that only contains relevant entries, which was used for this paper. Further details can be found in Appendix~\ref{sec:app_hyperparameters}.

\section{Results}

\subsection{Improved Performance and Efficiency}

The test results for our best model configurations and the benchmark models can be found in Table~\ref{tab:best_test_accuracies}. The best performing model is the fine-tuned BERT with single-step subgraphs. The fine-tuned BERT without any subgraphs were able to achieve slightly higher performance than the one from \citet{kim2023factkg}, which we suggest is due to finding better hyperparameters.

Additionally, our models were much faster to train. While the GEAR model used 2-3 days to train on an RTX 3090 GPU (reported by the authors by email), our QA-GNN only used 1.5 hours. The training time of our fine-tuned BERT model was greatly influenced by the size of the subgraphs we used. With no subgraphs, it took about 2 hours to train, while with the one-hop subgraph it took 10 hours. FactGenius was reported to use substantially more computational resources, running the LLM inference on a A100 GPU with 80GB VRAM for 8 hours.

\subsection{Successful Subgraphs Retrievals}

We now look at the different configurations for the subgraph retrievals, which greatly influenced the performance of the models. Since the \emph{direct} and \emph{contextual} approach only includes subgraphs if a certain requirement is fulfilled, it will result in some of the claims having empty subgraphs. In the training and validation set, 49.0\% of the graphs were non-empty for the \emph{direct} approach, and 62.5\% were non-empty for the \emph{contextual} approach. The \emph{single-step} method resulted in vastly bigger subgraphs.

\begin{table*}[tb]
    \centering
    \resizebox{0.95\textwidth}{!}{%
    \begin{tabular}{|c|ccccc|c|c}
    \Xhline{1.5pt}
    \textbf{Model} & \textbf{One-hop} & \textbf{Conjunction} & \textbf{Existence} & \textbf{Multi-hop} & \textbf{Negation} & \textbf{Total} \\ \hline
    BERT (no subgraphs) & 67.71 & 67.48 & 62.51 & 73.28 & 64.23 & 68.99 \\
    \hline
    BERT (direct) & 80.24 & 83.30 & 59.05 & 77.62 & 74.58 & 79.64 \\
    BERT (contextual) & 81.20 & 84.45 & 61.05 & 77.04 & 77.40 & 80.25 \\
    BERT (single-step) & \textbf{97.40} & \textbf{97.51} & \textbf{97.31} & \textbf{80.32} & \textbf{92.54} & \textbf{93.49} \\
    \hline
    QA-GNN (direct) & 74.60 & 74.01 & 58.97 & 76.41 & 74.12 & 75.01 \\
    QA-GNN (contextual) & 76.58 & 69.94 & 84.68 & 74.58 & 80.75 & 76.12 \\
    QA-GNN (single-step) & 79.08 & 74.43 & 83.37 & 74.72 & 79.60 & 78.08 \\
    
    \Xhline{1.5pt}
    \end{tabular}
    }
    \caption{\textbf{Test-set accuracy for different subgraph retrieval methods on \textsc{FactKG}.} The \emph{direct} approach only includes knowledge triples where both nodes appear in the claim, the \emph{contextual} also includes edges appearing in the claim, and the \emph{single-step} includes all knowledge triples where at least one node appears in the claim. The QA-GNN models used the single-step subgraph if the direct or contextual is empty, while the BERT models did not.}
    \label{tab:subgraph_test_results}
\end{table*}

While the QA-GNN could handle the big subgraphs efficiently, the fine-tuned BERT was severely slowed down when the size of the subgraphs got bigger. Therefore, we substituted any empty subgraphs with the \emph{single-step} subgraph when using QA-GNN, but kept the empty graphs when using fine-tuned BERT. This means that some claims for the direct and contextual BERT models were predicted only using the bias in the language model and the claim.

\begin{table*}[tb]
    \centering
    \resizebox{0.98\textwidth}{!}{
    \begin{tabular}{|l|ccccc|c|}
        \hline
        Model & \textbf{One-hop} & \textbf{Conjunction} & \textbf{Existence} & \textbf{Multi-hop} & \textbf{Negation} & \textbf{Total} \\
        & P / R / F1 & P / R / F1 & P / R / F1 & P / R / F1 & P / R / F1 & P / R / F1 \\
        \hline
        BERT (no subgraphs) & 71.89 / 51.66 / 60.12 & 75.44 / 34.20 / 47.06 & 59.52 / 73.63 / 65.82 & 85.19 / 60.90 / 71.03 & 58.88 / 73.13 / 65.24 & 75.25 / 54.00 / 62.88 \\
        \hline
        QA-GNN (direct) & 76.19 / 67.04 / 71.32 & 80.11 / 51.22 / 62.49 & 56.19 / 74.10 / 63.91 & 80.04 / 74.80 / 77.33 & 70.97 / 73.80 / 72.36 & 77.21 / 69.01 / 72.88 \\
        QA-GNN (contextual) & 84.79 / 61.29 / 71.15 & 80.27 / 38.29 / 51.85 & 81.83 / 88.38 / 84.98 & 82.31 / 67.17 / 73.98 & 77.26 / 82.26 / 79.68 & 84.10 / 62.78 / 71.89 \\
        QA-GNN (single-step) & 82.51 / 70.55 / 76.06 & 78.89 / 53.95 / 64.08 & 79.69 / 88.70 / 83.95 & 78.44 / 73.09 / 75.67 & 77.06 / 79.10 / 78.07 & 81.41 / 71.19 / 75.96 \\
        \hline
        BERT (contextual) & 83.05 / 75.51 / 79.10 & 88.60 / 72.56 / 79.78 & 59.68 / 63.42 / 61.49 & 84.10 / 70.67 / 76.80 & 75.84 / 74.46 / 75.15 & 83.30 / 74.28 / 78.53 \\
        BERT (direct) & 83.89 / 71.86 / 77.41 & 88.69 / 69.32 / 77.82 & 58.97 / 54.16 / 56.46 & 83.38 / 72.91 / 77.80 & 69.99 / 78.11 / 73.82 & 83.76 / 72.12 / 77.51 \\
        BERT (single-step) & \textbf{96.27} / \textbf{98.29} / \textbf{97.27} & \textbf{96.06} / \textbf{98.13} / \textbf{97.09} & \textbf{96.45} / \textbf{98.12} / \textbf{97.28} & \textbf{85.31} / \textbf{76.59} / \textbf{80.72} & \textbf{92.01} / \textbf{91.71} / \textbf{91.86} & \textbf{93.75} / \textbf{92.79} / \textbf{93.27} \\
        \hline
    \end{tabular}
    }
    \caption{\textbf{Precision} (P), \textbf{Recall} (R), and \textbf{F1 scores} for different models and subgraph types on the test-set.}
    \label{tab:precision_recall}
\end{table*}

The results can be found in Table~\ref{tab:subgraph_test_results} and Table~\ref{tab:precision_recall}. We see a clear improvement in BERT when using the direct subgraphs over none, a small improvement when using the contextual subgraphs, and a big improvement when using the single-step method. The same is true for the QA-GNN, but the differences in performance are smaller. The models score the lowest on multi-hop claims.

Since we used non-trainable subgraph retrieval methods and a frozen BERT for embedding the nodes and edges in the subgraphs, we performed this processing before training the models. During training, the models used a lookup table to get the subgraphs and the word embeddings, which significantly decreased the training time. The retrieval of all the subgraphs took about 15 minutes, and the embedding of all the words appearing in them took about 1 hour. We also tried training a QA-GNN without frozen embeddings, but it ran so slow that we were not able to carry out the training with our available computational resources.

\subsection{Competitive ChatGPT Performance}

\begin{table}[tb]
    \centering
    \resizebox{0.45\textwidth}{!}{%
    \begin{tabular}{|l|c|}
        \hline
        Model & Accuracy (mean ± std) \\
        \hline
        ChatGPT 25 questions & 73.67 $\pm$ 0.5 \\
        \hline
        ChatGPT 50 questions & \textbf{76.33} $\pm$ 3.3 \\
        \hline
        ChatGPT 100 questions & 73.00 $\pm$ 1.4 \\
        \hline
    \end{tabular}
    }
    \caption{\textbf{Test-set accuracy for different configurations of ChatGPT prompting.} The metrics are averaged over three runs. The prompts included 25, 50 or 100 claims at a time, but the same claims were used in all of the configurations.}
    \label{tab:chatgpt}
\end{table}

The results for the ChatGPT prompting can be found in Table~\ref{tab:chatgpt}. The accuracy is substantially lower than from our best models, but better than the baselines using only the claims. The accuracy is fairly consistent over the three runs, and we do not see a big difference between the amount of questions asked at a time.

We started with an initial prompt asking for just the truth values for a list of claims, and updated it to also include some training examples and to ask for explanations. Several configurations of the prompt were tested, and it was also improved based on feedback from ChatGPT.

\begin{figure*}[tb]
\centering
\footnotesize
\fbox{\begin{minipage}{\dimexpr \textwidth-2\fboxsep-2\fboxrule\relax}
\textbf{Task:}

Determine the truth value (True or False) of the following claims based on information verifiable from Wikipedia, as represented in the DBpedia knowledge graph. Provide your answers without using real-time internet searches or code analysis, relying solely on your pre-trained knowledge.

\textbf{Instructions:}
\begin{itemize}
    \item You will evaluate the following claims, presented one per line.
    \item Base your answers solely on your knowledge as of your last training cut-off.
    \item Provide answers in Python list syntax for easy copying.
    \item Respond with \texttt{True} for verifiable claims, and \texttt{False} otherwise.
    \item Include a brief explanation for each answer, explaining your reasoning based on your pre-training.
    \item If the claim is vague or lacks specific information, please make an educated guess on whether it is likely to be True or False.

\end{itemize}

\textbf{Output Format:}
Format your responses as a list in Python. Each entry should be a tuple, formatted as (claim number, answer, explanation).

\textbf{Example Claims:}

1. The Atatürk Monument is located in Izmir, Turkey, where the capital is Ankara.

2. Yes, Eamonn Butler's alma mater is the University of Texas System!

3. I have heard 300 North LaSalle was completed in 2009.

4. The band Clinton Gregory created an album in the rock style.
...

\textbf{Example output:}

[

\qquad    (1, True, "The Atatürk Monument is indeed located in Izmir, and the capital of Turkey is Ankara."),

\qquad    (2, False, "Eamonn Butler did not attend the University of Texas System; he is a British author and economist whose educational background does not include this institution."),

\qquad    (3, True, "300 North LaSalle in Chicago was indeed completed in 2009."),

\qquad    (4, False, "Clinton Gregory is primarily known as a country music artist, not rock."),

    ...

]

\textbf{Here are the actual claims you should answer:}
\end{minipage}}
\caption{\textbf{Final prompt used to get truth values from ChatGPT 4o}. The actual questions are not included, but were in the format of the \textbf{Example Claims}. The \textbf{Example Claims} are from the training set, and the \textbf{Example Output} is copy pasted from an actual ChatGPT answer.}
\label{fig:chatgpt_prompt}
\end{figure*}

We saw the biggest improvement when we asked for a short explanation of the answers, instead of just the truth values. Without asking for explanations, the amount of answers were often longer or shorter than the amount of questions, but this never happened when explanations were included. We added numbers to the questions to further help with this issue. We also saw a slight improvement by formulating the prompt with bullet point lists and by including some example inputs and outputs from the training set. The final prompt can be found in Figure~\ref{fig:chatgpt_prompt}.

\section{Discussion}

We were able to train better and more efficient models by simplifying the subgraph retrieval methods, both by using a fine-tuned BERT and a slightly modified QA-GNN model. While the QA-GNN models trained the fastest, the fine-tuned BERT clearly performed the best, significantly outperforming the benchmark models. This suggests that the simple logical subgraph retrievals worked better than the complex trained approaches in previous work. We suggest that the performance gain in the claim-only benchmark was due to slightly better hyperparameters.

All of the models performed better the bigger the subgraphs were. This suggests that the model architectures are able to utilize the relevant parts of the subgraphs, without needing an advanced subgraph retrieval step. This is emphasized by our fine-tuned BERT model achieving a 93.49\% test set accuracy by only using the single-step subgraphs, while the GEAR model from \citet{kim2023factkg}, which trained two language models to perform graph retrieval, achieved a 77.65\% test-set accuracy.

When examining the precision and recall metrics in Table~\ref{tab:precision_recall}, we see that most of the models has a higher precision than recall, except for the best performing model, the single-step BERT. However, the single-step BERT does have a lower recall for the multi-hop claims, which it performs significantly worse on. Therefore, the models mostly has a higher precision than recall when their performance is not so good, suggesting they are slightly more likely to predict ``false'' on claims that they are not confident about.

A limitation of our subgraph retrieval methods is that they never include nodes that are more than one step away from an entity node, while the trained approach from \citet{kim2023factkg} is dynamic and may include more. This might make the hypothesis that the simple subgraph retrieval methods will perform worse on \emph{multi-hop} claims than the dynamically trained, however, we see the exact opposite behavior. The best BERT and QA-GNN models score 80.32\% and 74.72\% at the multi-hop claims respectively, while the dynamic GEAR model scores 68.84\%, even lower than the models not using the subgraphs at all. We do however see that the best performing BERT model clearly performs the worst on the multi-hop claims compared to the other claim types, indicating that even bigger subgraphs might be beneficial.

While the sample size for the ChatGPT metrics were small, it does indicate that non-fine-tuned LLMs can achieve adequate few-shot performance compared to a fine-tuned claim-only BERT. The performance does not seem to be substantially compromised when the amount of questions prompted increases, so with a bigger access to computational resources, it might be possible to prompt the full test-set at once. The removal of fine-tuning greatly improves the ease of use if one only needs to verify a few claims. While we are hesitant to make any conclusion with the small sample size, we believe that the results serve as an approximate benchmark of how difficult the dataset is.

\section{Conclusion and Future Work}

Our results show that with simple, yet efficient methods for subgraph retrieval, our models were able to improve fact verification with knowledge graphs with respect to both performance and efficiency. The fine-tuned BERT model with single-step subgraphs clearly achieves the best performance, while the QA-GNN models are more efficient to train.

This indicates that complex models can work well with simple subgraph retrieval methods. Since the single-step subgraphs mostly contain information not relevant for the claims, the models are themselves able to filter away irrelevant information, and complex subgraph retrieval methods may hence not be necessary for accurate fact verification. Additionally, since the best performing model performed the poorest with multi-hop claims, future research could explore simple subgraphs retrieval methods allowing for bigger depths than one. Additionally, future work should also be directed towards running similar experiments on other datasets.

We also encourage researchers that have access to bigger computational resources to further explore the performance of LLMs for fact verification. A core limitation of our ChatGPT prompting was the inability to use the full test-set, and we consider this crucial for further development. We also think it would be especially interesting to make LLM and KG hybrid models. Since our results indicate that simple single-step subgraph retrievals outperform more complex methods, a promising path of future research would be to simply use both the claims and the single-step subgraphs as input to the LLM. If possible, the LLM could also be fine-tuned on the dataset. We also encourage future work to create fully reproducible results with LLMs, which we were unable to do.

\section{Limitations}

Our experiments with ChatGPT were done on a small sample of test questions, with a model that was not possible to seed, and therefore is not reproducible. Due to the small sample size, we are not able to directly compare the performance to other approaches. The lack of reproducibility, which stems from the state-of-the-art model that was available to the author is not fully publicly available, makes it impossible for other researchers to completely verify our findings. Additionally, the process for creating prompts were not standardized, we simply tried different configurations based on our own experience with using LLMs until we could not increase the validation accuracy further. Due to these limitations, one should therefore be very hesitant to make any confident conclusions based on the experiments we performed with ChatGPT.

Because our intention was to specifically explore different language models' abilities of fact verification with knowledge graphs on the \textsc{FactKG} dataset, we did not conduct any experiments on other datasets. It is possible that our results will not be consistent with other datasets.

Additionally, our selection of models and hyperparameter settings could be more diverse. Due to computational constraints, we did not perform a grid search for hyperparameters, but tuned hyperparameters one by one. Which parameters we searched for were not decided in advance. A predefined grid search might lead to a fairer and more reproducible approach. We did not experiment with different orderings of the knowledge triples for the fine-tuned BERT models, which could influence the performance.

\bibliography{bibliography}

\begin{thebibliography}{34}
\providecommand{\natexlab}[1]{#1}

\bibitem[{Achiam et~al.(2023)Achiam, Adler, Agarwal, Ahmad, Akkaya, Aleman, Almeida, Altenschmidt, Altman, Anadkat et~al.}]{achiam2023gpt4}
Josh Achiam, Steven Adler, Sandhini Agarwal, Lama Ahmad, Ilge Akkaya, Florencia~Leoni Aleman, Diogo Almeida, Janko Altenschmidt, Sam Altman, Shyamal Anadkat, et~al. 2023.
\newblock Gpt-4 technical report.
\newblock \emph{arXiv preprint arXiv:2303.08774}.

\bibitem[{Bekoulis et~al.(2021)Bekoulis, Papagiannopoulou, and Deligiannis}]{bekoulis2021review_on_fact_extraction}
Giannis Bekoulis, Christina Papagiannopoulou, and Nikos Deligiannis. 2021.
\newblock A review on fact extraction and verification.
\newblock \emph{ACM Computing Surveys (CSUR)}, 55(1):1--35.

\bibitem[{Bird et~al.(2009)Bird, Klein, and Loper}]{bird2009nltk}
Steven Bird, Ewan Klein, and Edward Loper. 2009.
\newblock \emph{Natural language processing with Python: analyzing text with the natural language toolkit}.
\newblock " O'Reilly Media, Inc.".

\bibitem[{Chen et~al.(2022)Chen, Zhang, Guo, Fan, and Cheng}]{chen2022gere}
Jiangui Chen, Ruqing Zhang, Jiafeng Guo, Yixing Fan, and Xueqi Cheng. 2022.
\newblock Gere: Generative evidence retrieval for fact verification.
\newblock In \emph{Proceedings of the 45th International ACM SIGIR Conference on Research and Development in Information Retrieval}, pages 2184--2189.

\bibitem[{Chen et~al.(2019)Chen, Wang, Chen, Zhang, Wang, Li, Zhou, and Wang}]{chen2019tabfact}
Wenhu Chen, Hongmin Wang, Jianshu Chen, Yunkai Zhang, Hong Wang, Shiyang Li, Xiyou Zhou, and William~Yang Wang. 2019.
\newblock Tabfact: A large-scale dataset for table-based fact verification.
\newblock \emph{arXiv preprint arXiv:1909.02164}.

\bibitem[{Cohen et~al.(2011)Cohen, Li, Yang, and Yu}]{cohen2011computational_journalism}
S~Cohen, C~Li, J~Yang, and C~Yu. 2011.
\newblock Computational journalism: A call to arms to database researchers, 148-151.
\newblock In \emph{5th Biennial Conference on Innovative Data Systems Research, CIDR}.

\bibitem[{Devlin et~al.(2018)Devlin, Chang, Lee, and Toutanova}]{devlin2018bert}
Jacob Devlin, Ming-Wei Chang, Kenton Lee, and Kristina Toutanova. 2018.
\newblock Bert: Pre-training of deep bidirectional transformers for language understanding.
\newblock \emph{arXiv preprint arXiv:1810.04805}.

\bibitem[{Gautam(2024)}]{gautam2024factgenius}
Sushant Gautam. 2024.
\newblock Factgenius: Combining zero-shot prompting and fuzzy relation mining to improve fact verification with knowledge graphs.
\newblock \emph{arXiv preprint arXiv:2406.01311}.

\bibitem[{Hanselowski et~al.(2018)Hanselowski, PVS, Schiller, Caspelherr, Chaudhuri, Meyer, and Gurevych}]{hanselowski2018retrospective_fake_news}
Andreas Hanselowski, Avinesh PVS, Benjamin Schiller, Felix Caspelherr, Debanjan Chaudhuri, Christian~M Meyer, and Iryna Gurevych. 2018.
\newblock A retrospective analysis of the fake news challenge stance detection task.
\newblock \emph{arXiv preprint arXiv:1806.05180}.

\bibitem[{Harris et~al.(2020)Harris, Millman, Van Der~Walt, Gommers, Virtanen, Cournapeau, Wieser, Taylor, Berg, Smith et~al.}]{harris2020numpy}
Charles~R Harris, K~Jarrod Millman, St{\'e}fan~J Van Der~Walt, Ralf Gommers, Pauli Virtanen, David Cournapeau, Eric Wieser, Julian Taylor, Sebastian Berg, Nathaniel~J Smith, et~al. 2020.
\newblock Array programming with numpy.
\newblock \emph{Nature}, 585(7825):357--362.

\bibitem[{Hassan et~al.(2015)Hassan, Adair, Hamilton, Li, Tremayne, Yang, and Yu}]{hassan2015quest_to_automate_fact-checking}
Naeemul Hassan, Bill Adair, James~T Hamilton, Chengkai Li, Mark Tremayne, Jun Yang, and Cong Yu. 2015.
\newblock The quest to automate fact-checking.
\newblock In \emph{Proceedings of the 2015 computation+ journalism symposium}. Citeseer.

\bibitem[{Hidey et~al.(2020)Hidey, Chakrabarty, Alhindi, Varia, Krstovski, Diab, and Muresan}]{hidey2020deseption}
Christopher Hidey, Tuhin Chakrabarty, Tariq Alhindi, Siddharth Varia, Kriste Krstovski, Mona Diab, and Smaranda Muresan. 2020.
\newblock Deseption: Dual sequence prediction and adversarial examples for improved fact-checking.
\newblock \emph{arXiv preprint arXiv:2004.12864}.

\bibitem[{Honnibal and Montani(2017)}]{spacy2}
Matthew Honnibal and Ines Montani. 2017.
\newblock {spaCy 2}: Natural language understanding with {B}loom embeddings, convolutional neural networks and incremental parsing.
\newblock To appear.

\bibitem[{Ioffe and Szegedy(2015)}]{ioffe2015batch_norm}
Sergey Ioffe and Christian Szegedy. 2015.
\newblock Batch normalization: Accelerating deep network training by reducing internal covariate shift.
\newblock In \emph{International conference on machine learning}, pages 448--456. pmlr.

\bibitem[{Kim et~al.(2023)Kim, Park, Kwon, Jo, Thorne, and Choi}]{kim2023factkg}
Jiho Kim, Sungjin Park, Yeonsu Kwon, Yohan Jo, James Thorne, and Edward Choi. 2023.
\newblock Factkg: Fact verification via reasoning on knowledge graphs.
\newblock \emph{arXiv preprint arXiv:2305.06590}.

\bibitem[{Lehmann et~al.(2015)Lehmann, Isele, Jakob, Jentzsch, Kontokostas, Mendes, Hellmann, Morsey, Van~Kleef, Auer et~al.}]{lehmann2015dbpedia}
Jens Lehmann, Robert Isele, Max Jakob, Anja Jentzsch, Dimitris Kontokostas, Pablo~N Mendes, Sebastian Hellmann, Mohamed Morsey, Patrick Van~Kleef, S{\"o}ren Auer, et~al. 2015.
\newblock Dbpedia--a large-scale, multilingual knowledge base extracted from wikipedia.
\newblock \emph{Semantic web}, 6(2):167--195.

\bibitem[{Liu et~al.(2019)Liu, Ott, Goyal, Du, Joshi, Chen, Levy, Lewis, Zettlemoyer, and Stoyanov}]{liu2019roberta}
Yinhan Liu, Myle Ott, Naman Goyal, Jingfei Du, Mandar Joshi, Danqi Chen, Omer Levy, Mike Lewis, Luke Zettlemoyer, and Veselin Stoyanov. 2019.
\newblock Roberta: A robustly optimized bert pretraining approach.
\newblock \emph{arXiv preprint arXiv:1907.11692}.

\bibitem[{Loshchilov and Hutter(2017)}]{loshchilov2017decoupled_adamw}
Ilya Loshchilov and Frank Hutter. 2017.
\newblock Decoupled weight decay regularization.
\newblock \emph{arXiv preprint arXiv:1711.05101}.

\bibitem[{Mishra et~al.(2022)Mishra, Suryavardan, Bhaskar, Chopra, Reganti, Patwa, Das, Chakraborty, Sheth, Ekbal et~al.}]{mishra2022factify}
Shreyash Mishra, S~Suryavardan, Amrit Bhaskar, Parul Chopra, Aishwarya~N Reganti, Parth Patwa, Amitava Das, Tanmoy Chakraborty, Amit~P Sheth, Asif Ekbal, et~al. 2022.
\newblock Factify: A multi-modal fact verification dataset.
\newblock In \emph{DE-FACTIFY@ AAAI}.

\bibitem[{Nie et~al.(2019)Nie, Chen, and Bansal}]{nie2019combining}
Yixin Nie, Haonan Chen, and Mohit Bansal. 2019.
\newblock Combining fact extraction and verification with neural semantic matching networks.
\newblock In \emph{Proceedings of the AAAI conference on artificial intelligence}, volume~33, pages 6859--6866.

\bibitem[{{Open AI}(2024)}]{2024chatgpt4o}
{Open AI}. 2024.
\newblock Hello gpt 4o.
\newblock \url{https://openai.com/index/hello-gpt-4o/}, Accessed 30.05.2024.

\bibitem[{Park et~al.(2021)Park, Min, Kang, Zettlemoyer, and Hajishirzi}]{park2021faviq}
Jungsoo Park, Sewon Min, Jaewoo Kang, Luke Zettlemoyer, and Hannaneh Hajishirzi. 2021.
\newblock Faviq: Fact verification from information-seeking questions.
\newblock \emph{arXiv preprint arXiv:2107.02153}.

\bibitem[{Paszke et~al.(2019)Paszke, Gross, Massa, Lerer, Bradbury, Chanan, Killeen, Lin, Gimelshein, Antiga et~al.}]{paszke2019pytorch}
Adam Paszke, Sam Gross, Francisco Massa, Adam Lerer, James Bradbury, Gregory Chanan, Trevor Killeen, Zeming Lin, Natalia Gimelshein, Luca Antiga, et~al. 2019.
\newblock Pytorch: An imperative style, high-performance deep learning library.
\newblock \emph{Advances in neural information processing systems}, 32.

\bibitem[{Schuster et~al.(2021)Schuster, Fisch, and Barzilay}]{schuster2021get_your_vitamin_C}
Tal Schuster, Adam Fisch, and Regina Barzilay. 2021.
\newblock Get your vitamin c! robust fact verification with contrastive evidence.
\newblock \emph{arXiv preprint arXiv:2103.08541}.

\bibitem[{Schuster et~al.(2019)Schuster, Shah, Yeo, Filizzola, Santus, and Barzilay}]{schuster2019towards_debiasing}
Tal Schuster, Darsh~J Shah, Yun Jie~Serene Yeo, Daniel Filizzola, Enrico Santus, and Regina Barzilay. 2019.
\newblock Towards debiasing fact verification models.
\newblock \emph{arXiv preprint arXiv:1908.05267}.

\bibitem[{Srivastava et~al.(2014)Srivastava, Hinton, Krizhevsky, Sutskever, and Salakhutdinov}]{srivastava2014dropout}
Nitish Srivastava, Geoffrey Hinton, Alex Krizhevsky, Ilya Sutskever, and Ruslan Salakhutdinov. 2014.
\newblock Dropout: a simple way to prevent neural networks from overfitting.
\newblock \emph{The journal of machine learning research}, 15(1):1929--1958.

\bibitem[{Thorne and Vlachos(2018)}]{thorne2018automated_fact_checking}
James Thorne and Andreas Vlachos. 2018.
\newblock Automated fact checking: Task formulations, methods and future directions.
\newblock \emph{arXiv preprint arXiv:1806.07687}.

\bibitem[{Thorne et~al.(2018)Thorne, Vlachos, Christodoulopoulos, and Mittal}]{thorne2018fever}
James Thorne, Andreas Vlachos, Christos Christodoulopoulos, and Arpit Mittal. 2018.
\newblock Fever: a large-scale dataset for fact extraction and verification.
\newblock \emph{arXiv preprint arXiv:1803.05355}.

\bibitem[{University Centre~for Information~Technology(2023)}]{2023uio_mlnodes}
University Of~Oslo University Centre~for Information~Technology. 2023.
\newblock Machine learning infrastructure (ml nodes).

\bibitem[{Veli{\v{c}}kovi{\'c} et~al.(2017)Veli{\v{c}}kovi{\'c}, Cucurull, Casanova, Romero, Lio, and Bengio}]{velivckovic2017graph_attention}
Petar Veli{\v{c}}kovi{\'c}, Guillem Cucurull, Arantxa Casanova, Adriana Romero, Pietro Lio, and Yoshua Bengio. 2017.
\newblock Graph attention networks.
\newblock \emph{arXiv preprint arXiv:1710.10903}.

\bibitem[{Wolf et~al.(2020)Wolf, Debut, Sanh, Chaumond, Delangue, Moi, Cistac, Rault, Louf, Funtowicz et~al.}]{wolf2020transformers}
Thomas Wolf, Lysandre Debut, Victor Sanh, Julien Chaumond, Clement Delangue, Anthony Moi, Pierric Cistac, Tim Rault, R{\'e}mi Louf, Morgan Funtowicz, et~al. 2020.
\newblock Transformers: State-of-the-art natural language processing.
\newblock In \emph{Proceedings of the 2020 conference on empirical methods in natural language processing: system demonstrations}, pages 38--45.

\bibitem[{Yasunaga et~al.(2021)Yasunaga, Ren, Bosselut, Liang, and Leskovec}]{yasunaga2021qa_gnn}
Michihiro Yasunaga, Hongyu Ren, Antoine Bosselut, Percy Liang, and Jure Leskovec. 2021.
\newblock Qa-gnn: Reasoning with language models and knowledge graphs for question answering.
\newblock \emph{arXiv preprint arXiv:2104.06378}.

\bibitem[{Zhou et~al.(2019)Zhou, Han, Yang, Liu, Wang, Li, and Sun}]{zhou2019gear}
Jie Zhou, Xu~Han, Cheng Yang, Zhiyuan Liu, Lifeng Wang, Changcheng Li, and Maosong Sun. 2019.
\newblock Gear: Graph-based evidence aggregating and reasoning for fact verification.
\newblock \emph{arXiv preprint arXiv:1908.01843}.

\bibitem[{Zlatkova et~al.(2019)Zlatkova, Nakov, and Koychev}]{zlatkova2019fact-checking_meets_fauxtography}
Dimitrina Zlatkova, Preslav Nakov, and Ivan Koychev. 2019.
\newblock Fact-checking meets fauxtography: Verifying claims about images.
\newblock \emph{arXiv preprint arXiv:1908.11722}.

\end{thebibliography}

\appendix

\section{Hyperparameter Details}
\label{sec:app_hyperparameters}

We used an AdamW optimizer \citep{loshchilov2017decoupled_adamw} and a linear learning rate scheduler with 50 warm up steps. We used the model from the epoch with lowest accuracy loss. The hyperparameters were tuned in a line search, first testing different learning rates, and testing all the other hyperparameters with the best learning rate. We searched for learning rates in $\{1e-3, 5e-4, 1e-4, 5e-5, 1e-5\}$ for all models. We initially set the batch size to 32, except for the BERT models with large subgraphs, which were set to 4 due to memory constraints. After finding the learning rate, we tried batch sizes in $\{32, 64, 128, 256\}$. For the QA-GNN model, we initially set the GNN dropout and the classifier dropout to 0.3, and tried values in $\{0, 0.1, 0.3, 0.5, 0.6\}$. We also tried to freeze some of the layers in the base model, and to use a RoBERTa \citep{liu2019roberta} instead of BERT \citep{devlin2018bert}, but neither of these approaches improved the validation loss.

\begin{table}[tb]
    \centering
    \footnotesize
    \resizebox{0.5\textwidth}{!}{%
    \begin{tabular}{|c|ccc|}
    \Xhline{1.5pt}

    \textbf{Model} & Learning Rate & Batch Size & Best Epoch \\ 
    \hline
    BERT (no subgraphs) & 1e-4 & 32 & 6 \\
    BERT (direct) & 1e-4 & 32 & 7  \\
    BERT (contextual) & 5e-5 & 8 & 7 \\
    BERT (single-step) & 5e-5 & 4 & 7  \\
    QA-GNN (direct) & 1e-4 & 128 & 8  \\
    QA-GNN (contextual) & 5e-5 & 64 & 17 \\
    QA-GNN (single-step) & 1e-5 & 128 & 20 \\

    \Xhline{1.5pt}
    \end{tabular}
    }
    \caption{\textbf{Final hyperparameters for the different models.} The direct QA-GNN model used GNN and classifier dropout rates of 0.3 and 0.3, while both the two other QA-GNN used 0.1 and 0.5.}
    \label{tab:hyperparameter_values}
\end{table}

The final hyperparameters can be found in Table~\ref{sec:app_hyperparameters}.

\section{Scientific Artifacts}
\label{sec:app_artifacts}

We conducted the experiments using several python libraries, including PyTorch version 2.0.1 \citep{paszke2019pytorch} with CUDA version 11.7, HuggingFace Transformers \citep{wolf2020transformers}, NumPy \citep{harris2020numpy}, SpaCy \citep{spacy2} and NLTK \citep{bird2009nltk}. We will make all the code used for this paper publicly available.

\end{document}